\documentclass{article}
\usepackage[preprint]{spconf}
\usepackage{spconf,amsmath,graphicx}

\usepackage{multirow}
\usepackage{algorithm}
\usepackage{algpseudocode}
\usepackage[table,xcdraw]{xcolor}
\usepackage[table]{xcolor}
\usepackage{amsmath}
\usepackage[colorlinks=true, allcolors=teal]{hyperref}

\usepackage{bm}

\usepackage{amssymb}

\DeclareMathOperator*{\argminA}{arg\,min}

\title{Segment Any Object Model (SAOM): Real-to-Simulation Fine-Tuning Strategy for Multi-Class Multi-Instance Segmentation}

\threeauthors
  {Mariia Khan, \\Jumana Abu-Khalaf, \\David Suter}
	{School of Science\\
	ECU, Australia\\
	mariia.khan@ecu.edu.au}
  {Yue Qiu}
	{AIRC\\
	AIST, Japan\\
	qiu.yue@aist.go.jp}
   {Yuren Cong, \\Bodo Rosenhahn}
	{TNT\\
	LUH, Germany\\
	cong@tnt.uni-hannover.de}

\begin{document}

\ninept
\maketitle
\begin{abstract}
Multi-class multi-instance segmentation is the task of identifying masks for multiple object classes and multiple instances of the same class within an image. The foundational Segment Anything Model (SAM) is designed for promptable multi-class multi-instance segmentation but tends to output part or sub-part masks in the ``everything'' mode for various real-world applications. Whole object segmentation masks play a crucial role for indoor scene understanding, especially in robotics applications. We propose a new domain invariant Real-to-Simulation (Real-Sim) fine-tuning strategy for SAM. We use object images and ground truth data collected from Ai2Thor simulator during fine-tuning (real-to-sim). To allow our Segment Any Object Model (SAOM) to work in the ``everything'' mode, we propose the novel \textit{nearest neighbour assignment} method, updating point embeddings for each ground-truth mask. SAOM is evaluated on our own dataset collected from Ai2Thor simulator. SAOM significantly improves on SAM, with a 28\% increase in mIoU and a 25\% increase in mAcc for 54 frequently-seen indoor object classes. Moreover, our Real-to-Simulation fine-tuning strategy demonstrates promising generalization performance in real environments without being trained on the real-world data (sim-to-real). The dataset and the code will be released after publication.
\end{abstract}

\section{INTRODUCTION}
\label{sec:intro}
\copyrightnotice{\copyright\ This work has been submitted to the IEEE for possible publication. Copyright may be transferred without notice, after which this version may no longer be accessible.}

Segmentation is one of the main tasks in computer vision, which is commonly used in applications such as object detection and tracking \cite{wang2019fast}, medical image analysis \cite{ma2023segment}, and robotics \cite{ehsani2018segan}. In the last few years, there has been a revolution in computer vision research due to large foundational models \cite{kirillov2023segment, brown2020language, touvron2023llama, zou2023segment, wang2023seggpt}, that can perform zero-shot generalization for new domains without additional training. 

The recent Segment Anything Model (SAM) \cite{kirillov2023segment} can use various input prompts like points, bounding boxes, or masks for segmentation. Moreover, SAM can work in the ``everything'' mode, where it is asked to provide a valid mask at each point in a pre-defined point grid on an image. For each point in the grid, SAM predicts multiple output masks. The nested masks are often at most three deep: a part,


\begin{figure}[t]
   \centering
  \includegraphics[width=0.98\linewidth]{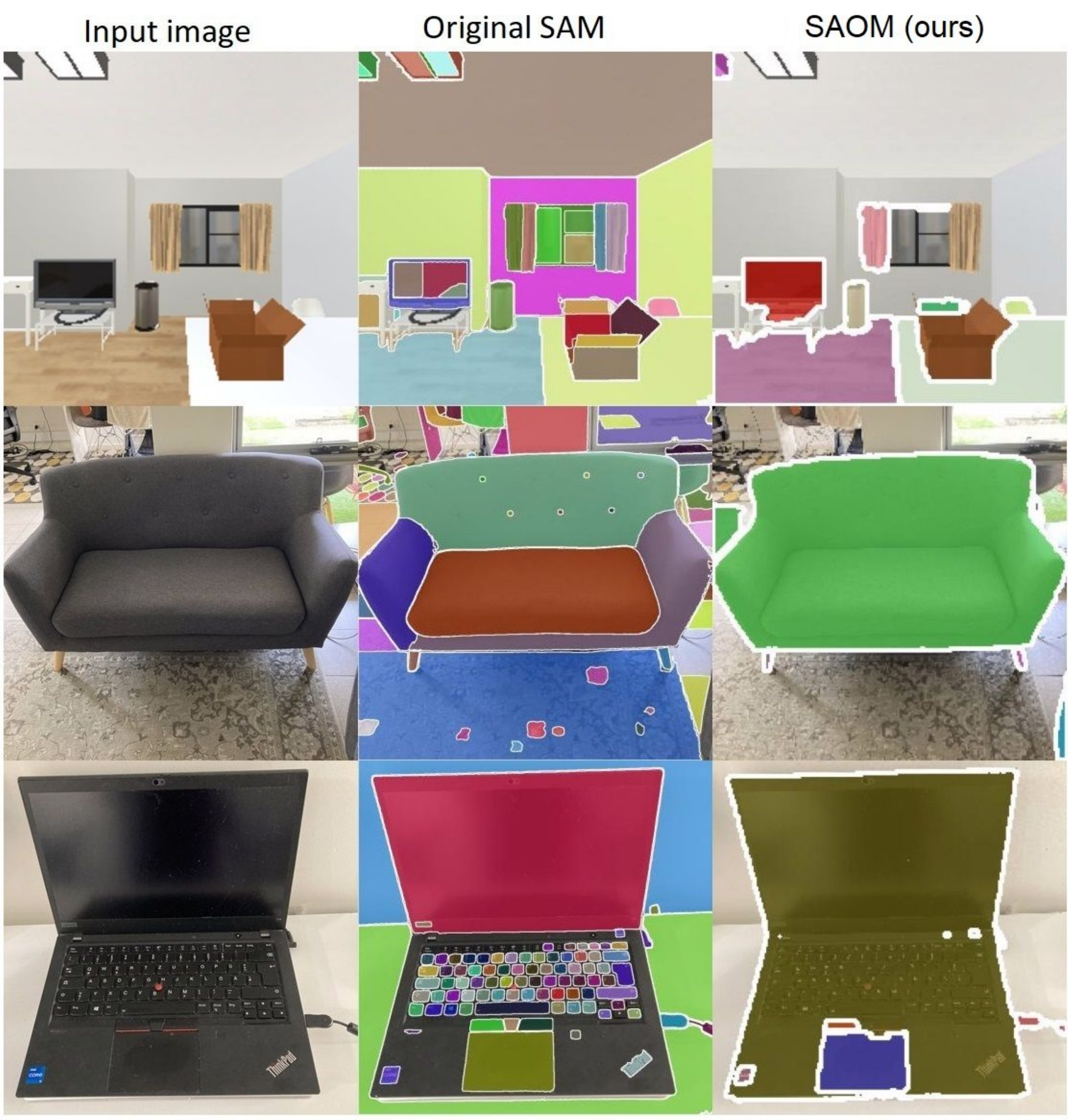}
   \caption{Comparison between vanilla SAM and our SAOM on images from Ai2Thor simulator (first row) and real-life scenes (second and third rows), where we apply the ``everything'' mode to obtain the displayed segmentation. We opted for thicker border lines in our SAOM model to emphasize the whole object nature of the segmentation masks. }
   \label{fig:sam13}
\vspace{-5mm}
\end{figure}

According to \cite{ji2023segment}, SAM can generalize well to natural images, but it has several limitations in real-world applications. SAM tends to favor selecting foreground object masks and is less effective in low-contrast applications \cite{ji2023segment}. SAM has also a limited understanding of professional data in the box mode and the ``everything'' mode \cite{ji2023segment}.

We examine the performance of SAM across several real-world segmentation applications in the ``everything'' mode and observe that SAM tends to output part and sub-part segmentation masks rather than segmentation masks of the complete objects. However, obtaining segmentation masks of the complete objects plays a crucial role in various computer vision applications: security for identification and tracking of objects, medical imaging for surgery planning and diagnosis, sports analytics for player tracking, and robotics for effective manipulation and grasping. Since these tasks involve analyzing previously unseen data, there are no predefined prompts for SAM input. Thus, SAM can be only used in the ``everything'' mode. 

To the best of our knowledge, there has been limited analysis in the literature regarding the applicability of SAM to indoor scene understanding \cite{huang2023instruct2act}, for which segmentation plays a key role. This is particularly relevant in applications like domestic robots, as the robotic agents learn from interactions with their surroundings. As agents usually are required to explore unseen environments, there are no prompts to serve as inputs to SAM, restricting its use to the "everything" mode. 

As collecting training data in real-world is costly and time-consuming, research related to various robotic tasks is typically performed using simulators. Our experiments, examining SAM’s performance in the ``everything'' mode, were conducted on images collected within the Ai2Thor \cite{kolve2017ai2} simulator (Fig. \ref{fig:sam11}). 

For indoor environments only a few simulators render realistic outputs \cite{savva2019habitat, gan2020threedworld}; most of the existing simulators \cite{kolve2017ai2, yan2018chalet, gao2019vrkitchen, li2021igibson} produce animated scenes. SAM was trained on the SA-1B dataset \cite{kirillov2023segment}, which consists of natural images from real-life scenes. It can be the main reason for SAM's tendency to output part or sub-part segmentation masks in the ``everything'' mode (Fig. \ref{fig:sam13}).

\begin{figure}[t]
   \centering

    \includegraphics[width=0.98\linewidth]{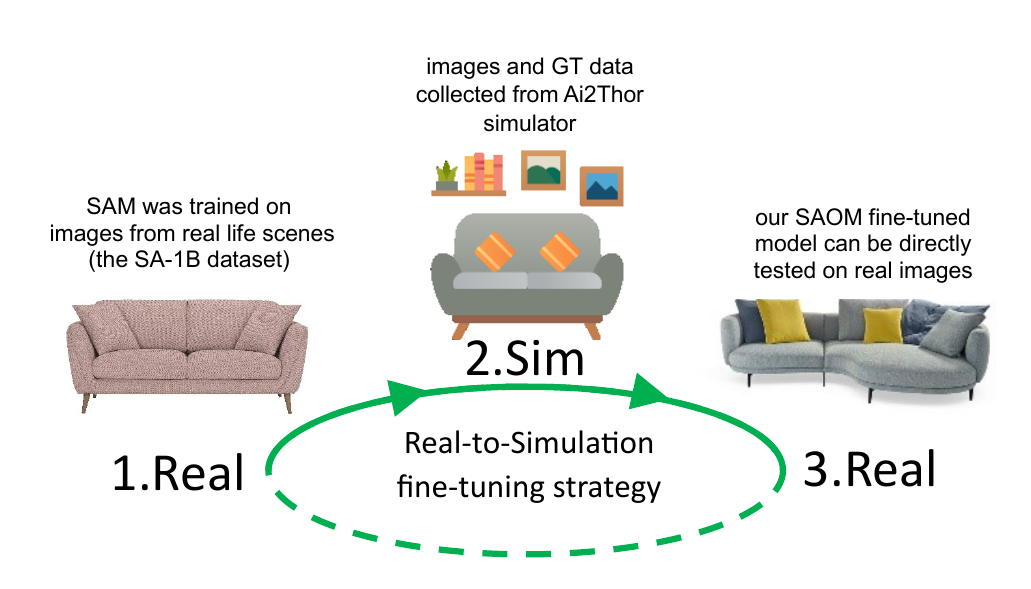}
   
   \caption{Our domain invariant Real-to-Simulation (Real-Sim) SAM's fine-tuning strategy. We use object images and GT data collected from Ai2Thor simulator during fine-tuning (real-to-sim) stage. The fine-tuned model, SAOM, can be directly tested on real images (sim-to-real) without being previously trained on real-world data.}
   \label{fig:sam1}
\vspace{-3mm}
\end{figure}

Therefore, we propose a new domain invariant fine-tuning strategy for multi-class multi-instance segmentation and apply it to the indoor scene understanding task. Inspired by works using synthetic data for training a model and then testing its performance on real-image benchmarks \cite{black2023bedlam, liu2020real}, we propose a novel Real-Sim fine-tuning method (Fig. \ref{fig:sam1}). In the real-to-sim training phase, images from simplified simulated environments are used to fine-tune SAM. The fine-tuning aims to recognize the whole objects instead of their parts in the ``everything'' mode. As SAM predicts object masks only and does not generate labels, we also add a classifier while fine-tuning. In the sim-to-real inference phase, the fine-tuned SAM is directly used in real-world scenarios without any training on real-world data. 

We evaluate our SAM’s fine-tuned version – Segment Any Object Model (SAOM) – adapted for semantic segmentation of objects in indoor environments (Fig. \ref{fig:sam13}) using Ai2Thor simulator \cite{kolve2017ai2}. We use 54 different interactable object classes of 3 sizes: small (e.g. mug, plate), medium (e.g. laptop, box), and large (e.g. dining table, sofa). Experimental results on sim-to-real transfer (Section \ref{sec:realdataset}) demonstrate that our Real-Sim strategy has a promising generalization performance in real environments (Fig. \ref{fig:sam13}) and low training costs. Our contributions can be summarized as follows:
\begin{itemize}
\item A new domain invariant SAM’s Real-Sim fine-tuning strategy (Section \ref{sec:method}). The proposed strategy is suitable for multi-class multi-instance semantic segmentation task in the ``everything'' mode. In the real-to-sim training phase, we propose a novel \textit{nearest neighbour assignment} method (Section \ref{sec:novelty}). We substitute the original object point prompts with their nearest neighbours from a pre-defined point grid on an image to make the model functional in the ``everything'' mode.
\item A SAM’s fine-tuned version adapted for semantic segmentation of whole objects in indoor environments – SAOM. The model achieves acceptable generalization results on real-world scenes without using real-life data during training (Section \ref{sec:experiments}). 
\item A new dataset for SAOM evaluation is created within the Ai2Thor simulator (Section \ref{sec:dataset}), which is also suitable for other segmentation tasks such as semantic or instance segmentation. Additionally, during the sim-to-real inference stage, we created and utilized a compact set of images in diverse contexts from real indoor environments (Section \ref{sec:realdataset}). Both dataset and code will be released after publication.

\end{itemize}

\section{RELATED WORK}

\begin{figure*}[t]
   \centering
   
     \includegraphics[width=0.88\linewidth]{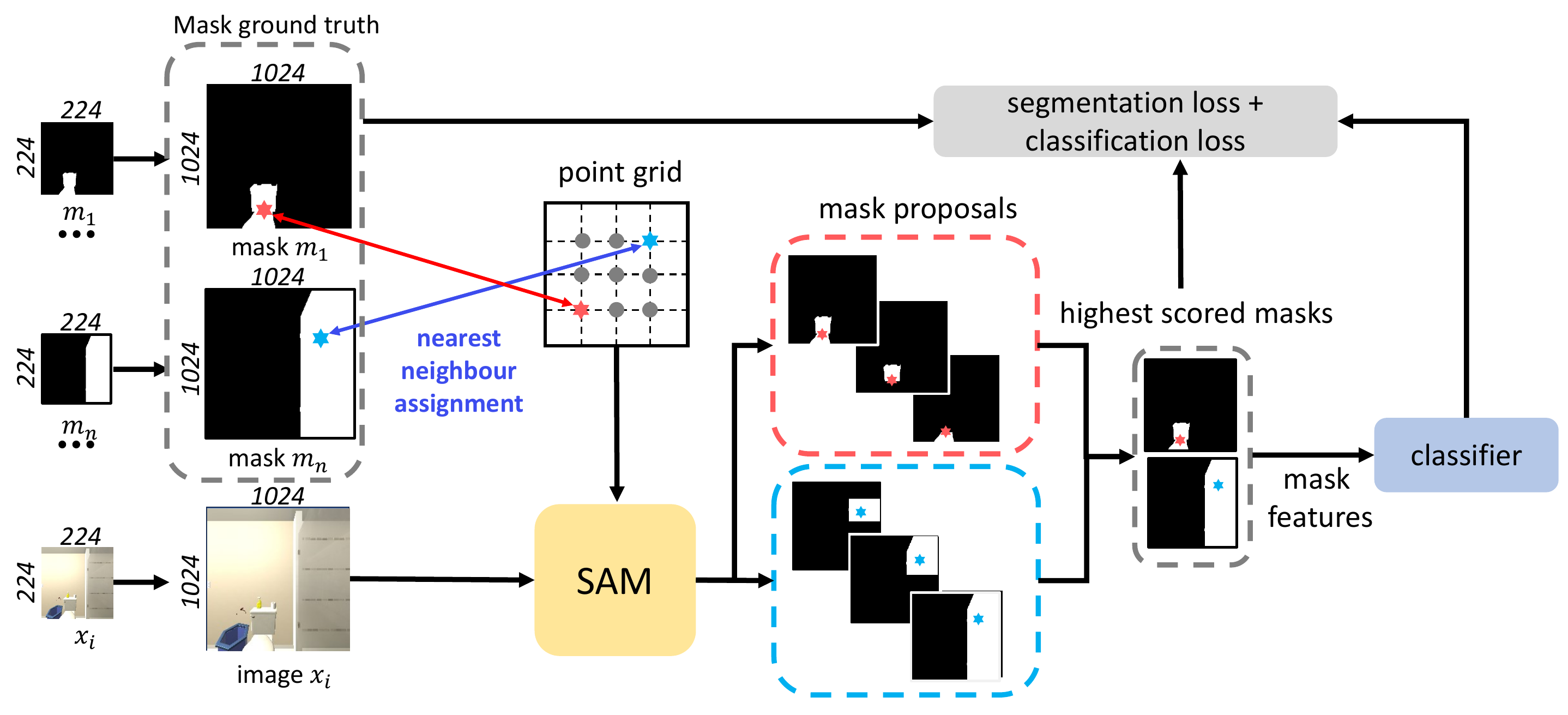}
   
   \caption{Our domain invariant SAM's fine-tuning strategy - SAOM - designed specifically for multi-class multi-instance semantic segmentation in the ``everything'' mode. We used our novel \textit{nearest neighbour assignment} method and substituted the original object point prompts with their nearest neighbours from a pre-defined point grid on an image to make the model functional in the ``everything'' mode.}
   \label{fig:overviewESAM}
\vspace{-3mm}
\end{figure*}
\subsection{SAM Applications}

A large number of research studies explored the generalization abilities of SAM and its applicability to various applications such as medical image analysis \cite{ma2023segment}, image editing \cite{xie2023edit}, camouflaged object detection \cite{tang2023can}, mirror or transparent objects detection \cite{han2023segment}, image captioning \cite{wang2023caption}, audio-visual localization \cite{mo2023av} and 3D reconstruction \cite{shen2023anything}. As far as we know, there has been limited analysis in the literature regarding the applicability of SAM to robotic tasks. In \cite{huang2023instruct2act} SAM is used to locate various small objects and use the output masks for the simple tabletop rearrangements. In \cite{liu2024online} SAM is used for robot navigation and manipulation in real-life indoor and outdoor scenes. Our research, presented in this paper, is the first to analyze SAM’s performance for the complex simulated indoor environments and its applicability to indoor scene understanding task.

\subsection{SAM's Fine-Tuning Strategies}

As SAM requires further improvement in some real-life applications, recent works have focused on developing fine-tuning strategies to enhance SAM's generalization abilities. Some models are domain-specific: e.g. SAM-Path \cite{zhang2023sam} and Ladder \cite{chai2023ladder} for medical domain or Segment Salient Object Model \cite{cui2023adaptive} for salient object detection. 

Other models are not designed to work in the ``everything'' mode, instead, they support specific input prompts for fine-tuning strategies: e.g. bounding boxes in AquaSAM \cite{xu2023aquasam}.  Thus, these models are not suitable for tasks with no prior knowledge about the environment. PerSAM \cite{zhang2023personalize} is used for the single-object segmentation task (with a single object within an image) in the ``everything'' mode, while we are interested in objects of multiple classes. Semantic-SAM \cite{li2023semantic} can be used for both automatic (the ``everything'' mode) and interactive mask generation (with point prompts). In the ``everything'' mode Semantic-SAM outputs controllable granularity masks from semantic, instance to part level when using different granularity prompts \cite{li2023semantic}. Note, that \cite{li2023semantic, zhang2023personalize, zhang2023sam, xu2023aquasam, chai2023ladder, cui2023adaptive, li2023semantic} only produce segmentation masks without labels and require a specific data format which is not easy to obtain from simulators. Open-Vocabulary SAM \cite{yuan2024open} is able to provide object labels together with segmentation masks, but it is not designed for use in the ``everything'' mode.

Thus, to the best of our knowledge, our Real-Sim
method is the first domain invariant SAM fine-tuning strategy designed specifically for multi-class multi-instance semantic segmentation in the ``everything'' mode. The model can automatically mask and label all objects of interest in an image without any manual prompt inputs. 

\section{DATASET}
\label{sec:dataset}

For the real-to-sim training stage, we collected a new segmentation dataset from the Ai2Thor simulator. For the initial scene setup, we created a dataset similar to RoomR \cite{weihs2021visual} with the agent's starting position and rotation, as well as the positions and rotations of the objects. We extend the RoomR dataset to include a larger amount of distinct rearrangement settings involving 54 different object types in 120 scenes. 

After the scene initialization, we collected 4 RGB image frames collected by the agent rotating in a scene. We use the action RotateRight and a 90° rotation angle to move the agent. For each image frame, corresponding binary object masks and labels were collected, together with a simple metadata dictionary mapping the image and its masks with labels. This data format can be extracted from most simulators, ensuring the domain-invariant nature of SAOM.

Overall, we provide a total of 293654 object masks for 33638 images collected from Ai2Thor. In total, 54 object classes were chosen for SAM's fine-tuning: 11 openable, 25 pickupable objects, and 18 receptacles, in or on which pickupable objects can be placed. Tables \ref{tab:real-to-sim} and \ref{tab:stat} provide the size and the statistics of our dataset, respectively. Note that objects were split into small, medium, and large sizes according to their visual appearance in the Ai2Thor simulator.

\begin{table}[tp!]
\centering
\caption{Size of Real-to-Sim Dataset}
\label{tab:real-to-sim}
\scalebox{0.90}{
\begin{tabular}{|l|l|l|l|
>{\columncolor[HTML]{C0C0C0}}l |}
\hline
                       & \cellcolor[HTML]{CBCEFB}TEST & \cellcolor[HTML]{CBCEFB}VAL & \cellcolor[HTML]{CBCEFB}TRAIN & \cellcolor[HTML]{CBCEFB}Total \\ \hline
Number of images       & 5660                         & 6500                        & 21478                         & 33638                         \\ \hline
Number of object masks & 63598                        & 47577                       & 192762                        & 303937                        \\ \hline
\end{tabular}%
}
\vspace{-3mm}
\end{table}


\begin{table}[t]
\caption{Real-to-Sim Dataset Statistics}
\label{tab:stat}
\centering
\label{tab:my-table}
\scalebox{0.75}{
\begin{tabular}{|l|l|l|l|l|l|l|l|}
\hline
\cellcolor[HTML]{CBCEFB} &
  \cellcolor[HTML]{CBCEFB} &
  \cellcolor[HTML]{CBCEFB} &
  \cellcolor[HTML]{CBCEFB} &
  \cellcolor[HTML]{CBCEFB} &
  \cellcolor[HTML]{CBCEFB} &
  \cellcolor[HTML]{CBCEFB} &
  \cellcolor[HTML]{CBCEFB} \\
\cellcolor[HTML]{CBCEFB} &
  \cellcolor[HTML]{CBCEFB} &
  \cellcolor[HTML]{CBCEFB} &
  \cellcolor[HTML]{CBCEFB} &
  \cellcolor[HTML]{CBCEFB} &
  \cellcolor[HTML]{CBCEFB} &
  \cellcolor[HTML]{CBCEFB} &
  \cellcolor[HTML]{CBCEFB} \\
\multirow{-3}{*}{\cellcolor[HTML]{CBCEFB}Type} &
  \multirow{-3}{*}{\cellcolor[HTML]{CBCEFB}Size} &
  \multirow{-3}{*}{\cellcolor[HTML]{CBCEFB}Class} &
  \multirow{-3}{*}{\cellcolor[HTML]{CBCEFB}Total} &
  \multirow{-3}{*}{\cellcolor[HTML]{CBCEFB}Type} &
  \multirow{-3}{*}{\cellcolor[HTML]{CBCEFB}Size} &
  \multirow{-3}{*}{\cellcolor[HTML]{CBCEFB}Class} &
  \multirow{-3}{*}{\cellcolor[HTML]{CBCEFB}Total} \\ \hline
 &
   &
  CounterTop &
  16965 &
   &
   &
  Statue &
  5195 \\ \cline{3-4} \cline{7-8} 
 &
   &
  DiningTable &
  4948 &
   &
   &
  Laptop &
  4766 \\ \cline{3-4} \cline{7-8} 
 &
   &
  Sofa &
  4390 &
   &
   &
  Pan &
  3131 \\ \cline{3-4} \cline{7-8} 
 &
   &
  Bed &
  4388 &
   &
   &
  Pot &
  3023 \\ \cline{3-4} \cline{7-8} 
 &
   &
  Bathtub &
  3741 &
   &
   &
  Vase &
  2231 \\ \cline{3-4} \cline{7-8} 
 &
   &
  Desk &
  3050 &
   &
   &
  TissueBox &
  1775 \\ \cline{3-4} \cline{7-8} 
 &
   &
  BathtubBasin &
  1258 &
   &
   &
  BaseballBat &
  1366 \\ \cline{3-4} \cline{7-8} 
 &
  \multirow{-8}{*}{\rotatebox{90}{Large} }&
  TVStand &
  1156 &
   &
  \multirow{-8}{*}{\rotatebox{90}{Medium}} &
  WateringCan &
  949 \\ \cline{2-4} \cline{6-8} 
 &
   &
  Shelf &
  19200 &
   &
   &
  SoapBottle &
  5435 \\ \cline{3-4} \cline{7-8} 
 &
   &
  Sink &
  9676 &
   &
   &
  ToiletPaper &
  5237 \\ \cline{3-4} \cline{7-8} 
 &
   &
  GarbageCan &
  9529 &
   &
   &
  Bowl &
  3944 \\ \cline{3-4} \cline{7-8} 
 &
   &
  SideTable &
  7668 &
   &
   &
  Mug &
  3772 \\ \cline{3-4} \cline{7-8} 
 &
   &
  SinkBasin &
  6541 &
   &
   &
  Box &
  3546 \\ \cline{3-4} \cline{7-8} 
 &
   &
  ArmChair &
  3821 &
   &
   &
  Plate &
  3376 \\ \cline{3-4} \cline{7-8} 
 &
   &
  CoffeeTable &
  2596 &
   &
   &
  Plunger &
  2993 \\ \cline{3-4} \cline{7-8} 
 &
  \multirow{-8}{*}{\rotatebox{90}{Medium}} &
  Ottoman &
  312 &
   &
   &
  SprayBottle &
  2971 \\ \cline{2-4} \cline{7-8} 
 &
   &
  StoveBurner &
  12070 &
   &
   &
  ScrubBrush &
  2848 \\ \cline{3-4} \cline{7-8} 
\multirow{-18}{*}{\rotatebox{90}{Receptacles}} &
  \multirow{-2}{*}{\rotatebox{90}{Small}} &
  ToiletPaperHanger &
  3058 &
   &
   &
  Book &
  2751 \\ \cline{1-4} \cline{7-8} 
 &
   &
  Cabinet &
  43009 &
   &
   &
  DishSponge &
  2711 \\ \cline{3-4} \cline{7-8} 
 &
   &
  Drawer &
  38547 &
   &
   &
  Cup &
  2495 \\ \cline{3-4} \cline{7-8} 
 &
   &
  Toilet &
  5135 &
   &
   &
  SoapBar &
  2393 \\ \cline{3-4} \cline{7-8} 
 &
   &
  Blinds &
  4530 &
   &
   &
  AlarmClock &
  2051 \\ \cline{3-4} \cline{7-8} 
 &
   &
  Fridge &
  4170 &
   &
   &
  Newspaper &
  1358 \\ \cline{3-4} \cline{7-8} 
 &
   &
  Microwave &
  3490 &
   &
   &
  PaperTowelRoll &
  1259 \\ \cline{3-4} \cline{7-8} 
 &
   &
  Dresser &
  2812 &
  \multirow{-25}{*}{\rotatebox{90}{Pickupable}} &
  \multirow{-17}{*}{\rotatebox{90}{Small}} &
  BasketBall &
  879 \\ \cline{3-8} 
 &
   &
  ShowerDoor &
  2253 &
   &
   &
   &
   \\ \cline{3-4} 
 &
   &
  ShowerCurtain &
  1820 &
   &
   &
   &
   \\ \cline{3-4} 
 &
   &
  Safe &
  585 &
   &
   &
   &
   \\ \cline{3-4} 
\multirow{-11}{*}{\rotatebox{90}{Openable}} &
  \multirow{-11}{*}{\rotatebox{90}{Medium and Large}} &
  LaundryHumper &
  481 &
   &
   &
   &
   \\ \hline
\end{tabular}%
    }
\vspace{-3mm}
\end{table}

\section{METHOD}
\label{sec:method}

In this paper, we propose a novel fine-tuning method for the Segment Anything Model (SAM) ~\cite{kirillov2023segment} to facilitate multi-class multi-instance segmentation for the indoor scene understanding. Instead of collecting training data in the real world,  we use object images and ground truth data collected from a simulator during fine-tuning (real-to-sim). The fine-tuned model, SAOM, can be directly tested on real images during the sim-to-real stage.

The aim of the training stage is to correctly predict a whole-object segmentation mask and a class (label) for all objects of interest in a given image frame collected on the path of the robotic agent.

\subsection{Real-to-Sim Fine-tuning} 

Fig. \ref{fig:overviewESAM} presents the overall pipeline of our SAM fine-tuning strategy. The model expects an input image \(x_{i} \in \mathbb{R}^{H \times W \times C}\), a set of whole-object binary masks \(m_{1},...,m_{n} \in \mathbb{M}^{H \times W \times 1}\) and their labels \(l_{1},...,l_{n} \in \mathbb{L}\) for all objects of interest in that image, where \(H\), \(W\), and \(C\) denote height, width, and channel numbers, respectively. Note that images from simulators usually have low resolution (e.g. 224 × 224 pixels) because of memory and computational efficiency issues, while vanilla SAM was trained on high resolution images of 1024 × 1024 pixels. Thus, the input image and binary masks are first interpolated to achieve the 1024 × 1024 resolution. 

We extend PerSAM \cite{zhang2023personalize} for multi-class multi-instance segmentation by processing one pair of a ground truth mask \(m_{n}\) with a label \(l_{n}\) and an input image \(x_{i}\) at a time. As the mechanism of localizing visual targets by location priors in PerSAM \cite{zhang2023personalize} is not adapted to work in the ``everything'' mode, we propose a novel \textit{nearest neighbour assignment} method for the selection of target object prompts. We obtain a location prior of a target object \(n\) in an image \(x_{i}\) using SAM. Then, we substitute the original location priors with their nearest neighbours from a pre-defined point grid on that image in the ``everything'' mode. 

On top of SAM, we add an additional classifier layer to predict a label for the target object. As SAM's mask decoder has the capability to generate multiple results, the default number of outputs is set to three. We choose a segmentation mask with the highest calculated IoU score out of 3 provided by SAM. The fine-tuning loss consists of cross entropy, focal and dice loss. We repeat the above refinement process for all pairs of ground truth masks \(m_{1},...,m_{n} \in \mathbb{M}^{H \times W \times 1}\) and labels \(l_{1},...,l_{n} \in \mathbb{L}\) for a given image \(x_{i}\).

    \begin{algorithm}[b]
    \caption{Nearest Neighbour Assignment}
    \label{algo:nearestpointassignment}
    \begin{algorithmic}
    
    \State Define $\bm{P}=[p_1, p_2, ..., p_K]$; \Comment{Point grid}
    \State Calculate $F_x$ = $Enc (x_i)$; \Comment{Image features}
    \For{$m_i \gets 1$ to $m_n$}                    
        \State Get $F_m$ = $Enc (m_i)$; \Comment{Mask features}
        \State Build $S$ = $F_x$$F_m$; \Comment{Location confidence map}
        \State Choose $c_i$ = $argmax(S)$; \Comment{Object location prior}
        \State Find $p_i$ = $\argminA_{p \in \bm{P}}  \mathcal{D}(c_i, p)$; \Comment{Nearest neighbour}
        \State Assign $p_i$ to $m_i$; 
    \EndFor
    
    \end{algorithmic}
    \end{algorithm}
\subsection{Nearest Neighbour Assignment}
\label{sec:novelty}
The pre-trained point embeddings as input prompts in SAM are learned from a large-scale real-world dataset. They can easily result in under-segmentation or over-segmentation in a specific scene. To update these point embeddings, we introduce a \textit{nearest neighbour assignment} method that assigns a corresponding point to each ground-truth mask during the real-to-sim fine-tuning stage.

Given a ground-truth binary mask $m_i$, we only consider the positive region in the image where the object exists to derive object mask features $F_m$ within that region using SAM’s pre-trained image encoder $Enc$. We adopt an average pooling to aggregate its global visual embedding and calculate the cosine similarity $S$ between the $F_m$ and an input image feature $F_x$.  Thus, we build a location confidence map \cite{zhang2023personalize} of the target object within a given image \(x_{i}\). We select the pixel coordinate $c_i$ with the highest similarity value from the confidence map. The distances between the chosen pixel and all points  $\bm{P}=[p_1, p_2, ..., p_K]$ uniformly distributed on the 2D image (point grid) are then computed.
The nearest neighbour $p_i$ to the chosen pixel can be formulated as:
\begin{equation}
\centering
p_i = \argminA_{p \in \bm{P}}  \mathcal{D}(c_i, p)
\label{eq:NNB}
\end{equation}
where $\mathcal{D}$ indicates the euclidean distance. 
A unique prompt point $p_i$ from a pre-defined point grid is assigned automatically in this way for each segmentation mask \(m_{1},...,m_{n} \in \mathbb{M}^{H \times W \times 1}\). The details of the \textit{nearest neighbour assignment} method are in Algorithm \ref{algo:nearestpointassignment}.

\section{EXPERIMENTS}
\label{sec:experiments}
\subsection{Implementation Details}

In our experiments we used the Ai2Thor \cite{kolve2017ai2} simulator, as it offers a rich collection of various size objects. We collected RGB images and binary object masks with 224 × 224 resolution to reduce the computational complexity. We adopt pre-trained SAM with a ViT-B \cite{dosovitskiy2020image} image encoder to increase the training speed. SAOM was trained for 200 epochs using the 32 × 32 point grid for the ``everything'' mode. We set the initial learning rate as \(10^{-3} \) and adopt the AdamW \cite{loshchilov2017decoupled} optimizer with a cosine scheduler. We additionally tuned the following SAM automatic mask generator parameters: prediction IoU threshold (from 0.58 to 0.3), box IoU cutoff (from 0.4 to 0.3) and minimum mask region area (from 0 to 150). These improvements help to further filter duplicate masks and remove disconnected regions in masks. 

\subsection{Real-to-Sim Evaluation}

\begin{table}[t]
\caption{Evaluation results for pickupable and receptacle objects of different sizes}
\label{tab:pick}
\centering
\scalebox{0.48}{
\resizebox{\textwidth}{!}{%
\begin{tabular}{|cc|c|cc|
>{\columncolor[HTML]{67FD9A}}c 
>{\columncolor[HTML]{67FD9A}}c c|}
\hline
\multicolumn{2}{|c|}{\cellcolor[HTML]{CBCEFB}} &
  \cellcolor[HTML]{CBCEFB} &
  \multicolumn{2}{c|}{\cellcolor[HTML]{CBCEFB}Original SAM} &
  \multicolumn{3}{c|}{\cellcolor[HTML]{CBCEFB}SAOM after 200 epoch} \\ \cline{4-8} 
\multicolumn{2}{|c|}{\cellcolor[HTML]{CBCEFB}} &
  \cellcolor[HTML]{CBCEFB} &
  \multicolumn{1}{c|}{\cellcolor[HTML]{CBCEFB}} &
  \cellcolor[HTML]{CBCEFB} &
  \multicolumn{1}{c|}{\cellcolor[HTML]{CBCEFB}} &
  \multicolumn{1}{c|}{\cellcolor[HTML]{CBCEFB}} &
  \cellcolor[HTML]{CBCEFB} \\
\multicolumn{2}{|c|}{\multirow{-3}{*}{\cellcolor[HTML]{CBCEFB}\begin{tabular}[c]{@{}c@{}}Objects\\ of different sizes\end{tabular}}} &
  \multirow{-3}{*}{\cellcolor[HTML]{CBCEFB}Occurences} &
  \multicolumn{1}{c|}{\multirow{-2}{*}{\cellcolor[HTML]{CBCEFB}IoU}} &
  \multirow{-2}{*}{\cellcolor[HTML]{CBCEFB}Acc} &
  \multicolumn{1}{c|}{\multirow{-2}{*}{\cellcolor[HTML]{CBCEFB}IoU}} &
  \multicolumn{1}{c|}{\multirow{-2}{*}{\cellcolor[HTML]{CBCEFB}Acc}} &
  \multirow{-2}{*}{\cellcolor[HTML]{CBCEFB}\begin{tabular}[c]{@{}c@{}}Classification\\ Accuracy\end{tabular}} \\ \hline
\multicolumn{1}{|c|}{} &
  CounterTop &
  2151 &
  \multicolumn{1}{c|}{22} &
  64.08 &
  \multicolumn{1}{c|}{\cellcolor[HTML]{67FD9A}37.83} &
  \multicolumn{1}{c|}{\cellcolor[HTML]{FD6864}49.94} &
  0.28 \\ \cline{2-8} 
\multicolumn{1}{|c|}{} &
  DiningTable &
  960 &
  \multicolumn{1}{c|}{43.45} &
  72.75 &
  \multicolumn{1}{c|}{\cellcolor[HTML]{67FD9A}49.29} &
  \multicolumn{1}{c|}{\cellcolor[HTML]{FD6864}55.09} &
  0.41 \\ \cline{2-8} 
\multicolumn{1}{|c|}{} &
  SideTable &
  911 &
  \multicolumn{1}{c|}{3.38} &
  39.81 &
  \multicolumn{1}{c|}{\cellcolor[HTML]{67FD9A}24.83} &
  \multicolumn{1}{c|}{\cellcolor[HTML]{67FD9A}46.63} &
  0.28 \\ \cline{2-8} 
\multicolumn{1}{|c|}{} &
  Sofa &
  792 &
  \multicolumn{1}{c|}{25.44} &
  34.62 &
  \multicolumn{1}{c|}{\cellcolor[HTML]{67FD9A}49.19} &
  \multicolumn{1}{c|}{\cellcolor[HTML]{67FD9A}51.18} &
  0.55 \\ \cline{2-8} 
\multicolumn{1}{|c|}{} &
  ArmChair &
  690 &
  \multicolumn{1}{c|}{16.83} &
  42.69 &
  \multicolumn{1}{c|}{\cellcolor[HTML]{67FD9A}53.4} &
  \multicolumn{1}{c|}{\cellcolor[HTML]{67FD9A}59.81} &
  0.3 \\ \cline{2-8} 
\multicolumn{1}{|c|}{\multirow{-6}{*}{\rotatebox{90}{Large}}} &
  Bed &
  562 &
  \multicolumn{1}{c|}{52.3} &
  59.11 &
  \multicolumn{1}{c|}{\cellcolor[HTML]{67FD9A}60} &
  \multicolumn{1}{c|}{\cellcolor[HTML]{67FD9A}62.34} &
  0.43 \\ \hline
\multicolumn{1}{|c|}{} &
  Sink &
  2383 &
  \multicolumn{1}{c|}{9.01} &
  48.3 &
  \multicolumn{1}{c|}{\cellcolor[HTML]{67FD9A}22.67} &
  \multicolumn{1}{c|}{\cellcolor[HTML]{FD6864}44.19} &
  0.35 \\ \cline{2-8} 
\multicolumn{1}{|c|}{} &
  GarbageCan &
  1494 &
  \multicolumn{1}{c|}{5.96} &
  56.51 &
  \multicolumn{1}{c|}{\cellcolor[HTML]{67FD9A}79.86} &
  \multicolumn{1}{c|}{\cellcolor[HTML]{67FD9A}91.4} &
  0.67 \\ \cline{2-8} 
\multicolumn{1}{|c|}{} &
  SinkBasin &
  982 &
  \multicolumn{1}{c|}{7.52} &
  59.3 &
  \multicolumn{1}{c|}{\cellcolor[HTML]{67FD9A}17.86} &
  \multicolumn{1}{c|}{\cellcolor[HTML]{FD6864}52.73} &
  0.38 \\ \cline{2-8} 
\multicolumn{1}{|c|}{} &
  Laptop &
  710 &
  \multicolumn{1}{c|}{7.46} &
  42.64 &
  \multicolumn{1}{c|}{\cellcolor[HTML]{67FD9A}64.86} &
  \multicolumn{1}{c|}{\cellcolor[HTML]{67FD9A}80.37} &
  0.29 \\ \cline{2-8} 
\multicolumn{1}{|c|}{} &
  Box &
  639 &
  \multicolumn{1}{c|}{8.3} &
  48.35 &
  \multicolumn{1}{c|}{\cellcolor[HTML]{67FD9A}56.03} &
  \multicolumn{1}{c|}{\cellcolor[HTML]{67FD9A}69.12} &
  0.51 \\ \cline{2-8} 
\multicolumn{1}{|c|}{\multirow{-6}{*}{\rotatebox{90}{Medium}}} &
  Shelf &
  582 &
  \multicolumn{1}{c|}{5.7} &
  35.05 &
  \multicolumn{1}{c|}{\cellcolor[HTML]{67FD9A}15.26} &
  \multicolumn{1}{c|}{\cellcolor[HTML]{67FD9A}35.85} &
  0.25 \\ \hline
\multicolumn{1}{|c|}{} &
  TioletPaper &
  997 &
  \multicolumn{1}{c|}{0.26} &
  23.58 &
  \multicolumn{1}{c|}{\cellcolor[HTML]{67FD9A}29.58} &
  \multicolumn{1}{c|}{\cellcolor[HTML]{67FD9A}80.31} &
  0.29 \\ \cline{2-8} 
\multicolumn{1}{|c|}{} &
  SoapBottle &
  835 &
  \multicolumn{1}{c|}{1.05} &
  24.94 &
  \multicolumn{1}{c|}{\cellcolor[HTML]{67FD9A}51.75} &
  \multicolumn{1}{c|}{\cellcolor[HTML]{67FD9A}84.17} &
  0.32 \\ \cline{2-8} 
\multicolumn{1}{|c|}{} &
  Mug &
  631 &
  \multicolumn{1}{c|}{0.7} &
  28.51 &
  \multicolumn{1}{c|}{\cellcolor[HTML]{67FD9A}28.09} &
  \multicolumn{1}{c|}{\cellcolor[HTML]{67FD9A}94.07} &
  0.26 \\ \cline{2-8} 
\multicolumn{1}{|c|}{} &
  StoveBurner &
  551 &
  \multicolumn{1}{c|}{2.33} &
  47.18 &
  \multicolumn{1}{c|}{\cellcolor[HTML]{67FD9A}12.97} &
  \multicolumn{1}{c|}{\cellcolor[HTML]{67FD9A}67.51} &
  0.55 \\ \cline{2-8} 
\multicolumn{1}{|c|}{\multirow{-5}{*}{\rotatebox{90}{Small}}} &
  Plate &
  524 &
  \multicolumn{1}{c|}{2.26} &
  45.65 &
  \multicolumn{1}{c|}{\cellcolor[HTML]{67FD9A}49.98} &
  \multicolumn{1}{c|}{\cellcolor[HTML]{67FD9A}93.44} &
  0.21 \\ \hline
\end{tabular}%
}}
\vspace{-3mm}
\end{table}
\begin{table}[b]
\caption{Evaluation results for openable objects}
\label{tab:open}
\centering
\scalebox{0.48}{
\resizebox{\textwidth}{!}{%
\begin{tabular}{|cc|c|cc|
>{\columncolor[HTML]{9AFF99}}c cc|}
\hline
\multicolumn{2}{|c|}{\cellcolor[HTML]{CBCEFB}} &
  \cellcolor[HTML]{CBCEFB} &
  \multicolumn{2}{c|}{\cellcolor[HTML]{CBCEFB}Original SAM} &
  \multicolumn{3}{c|}{\cellcolor[HTML]{CBCEFB}SAOM after 200th epoch} \\ \cline{4-8} 
\multicolumn{2}{|c|}{\cellcolor[HTML]{CBCEFB}} &
  \cellcolor[HTML]{CBCEFB} &
  \multicolumn{1}{c|}{\cellcolor[HTML]{CBCEFB}} &
  \cellcolor[HTML]{CBCEFB} &
  \multicolumn{1}{c|}{\cellcolor[HTML]{CBCEFB}} &
  \multicolumn{1}{c|}{\cellcolor[HTML]{CBCEFB}} &
  \cellcolor[HTML]{CBCEFB} \\
\multicolumn{2}{|c|}{\multirow{-3}{*}{\cellcolor[HTML]{CBCEFB}Objects}} &
  \multirow{-3}{*}{\cellcolor[HTML]{CBCEFB}Occurences} &
  \multicolumn{1}{c|}{\multirow{-2}{*}{\cellcolor[HTML]{CBCEFB}IoU}} &
  \multirow{-2}{*}{\cellcolor[HTML]{CBCEFB}Acc} &
  \multicolumn{1}{c|}{\multirow{-2}{*}{\cellcolor[HTML]{CBCEFB}IoU}} &
  \multicolumn{1}{c|}{\multirow{-2}{*}{\cellcolor[HTML]{CBCEFB}Acc}} &
  \multirow{-2}{*}{\cellcolor[HTML]{CBCEFB}\begin{tabular}[c]{@{}c@{}}Classification \\ Accuracy\end{tabular}} \\ \hline
\multicolumn{1}{|c|}{} &
  Drawer &
  2377 &
  \multicolumn{1}{c|}{5.41} &
  59.7 &
  \multicolumn{1}{c|}{\cellcolor[HTML]{9AFF99}17.85} &
  \multicolumn{1}{c|}{\cellcolor[HTML]{FD6864}51.64} &
  0.41 \\ \cline{2-8} 
\multicolumn{1}{|c|}{} &
  Cabinet &
  1755 &
  \multicolumn{1}{c|}{14.87} &
  49.13 &
  \multicolumn{1}{c|}{\cellcolor[HTML]{9AFF99}22.38} &
  \multicolumn{1}{c|}{\cellcolor[HTML]{FD6864}36.36} &
  0.83 \\ \cline{2-8} 
\multicolumn{1}{|c|}{} &
  Toilet &
  1659 &
  \multicolumn{1}{c|}{5.42} &
  47.62 &
  \multicolumn{1}{c|}{\cellcolor[HTML]{9AFF99}60.73} &
  \multicolumn{1}{c|}{\cellcolor[HTML]{9AFF99}71.08} &
  0.61 \\ \cline{2-8} 
\multicolumn{1}{|c|}{} &
  Fridge &
  678 &
  \multicolumn{1}{c|}{72.62} &
  78.67 &
  \multicolumn{1}{c|}{\cellcolor[HTML]{9AFF99}76.82} &
  \multicolumn{1}{c|}{\cellcolor[HTML]{9AFF99}81.97} &
  0.67 \\ \cline{2-8} 
\multicolumn{1}{|c|}{} &
  Microwave &
  584 &
  \multicolumn{1}{c|}{23.18} &
  57.13 &
  \multicolumn{1}{c|}{\cellcolor[HTML]{9AFF99}46.86} &
  \multicolumn{1}{c|}{\cellcolor[HTML]{9AFF99}57.74} &
  0.63 \\ \cline{2-8} 
\multicolumn{1}{|c|}{} &
  Blinds &
  580 &
  \multicolumn{1}{c|}{12.31} &
  62.04 &
  \multicolumn{1}{c|}{\cellcolor[HTML]{9AFF99}75.7} &
  \multicolumn{1}{c|}{\cellcolor[HTML]{9AFF99}84.45} &
  0.95 \\ \cline{2-8} 
\multicolumn{1}{|c|}{} &
  Dresser &
  567 &
  \multicolumn{1}{c|}{25.73} &
  77.16 &
  \multicolumn{1}{c|}{\cellcolor[HTML]{9AFF99}53.93} &
  \multicolumn{1}{c|}{\cellcolor[HTML]{FD6864}65.05} &
  0.21 \\ \cline{2-8} 
\multicolumn{1}{|c|}{\multirow{-8}{*}{\rotatebox{90}{Medium and large}}} &
  ShowerDoor &
  526 &
  \multicolumn{1}{c|}{44.37} &
  52.17 &
  \multicolumn{1}{c|}{\cellcolor[HTML]{9AFF99}54.25} &
  \multicolumn{1}{c|}{\cellcolor[HTML]{9AFF99}62.22} &
  0.91 \\ \hline
\end{tabular}%
}}
\end{table}

\begin{figure*}[t]
   \centering
   
   \includegraphics[width=0.88\linewidth]{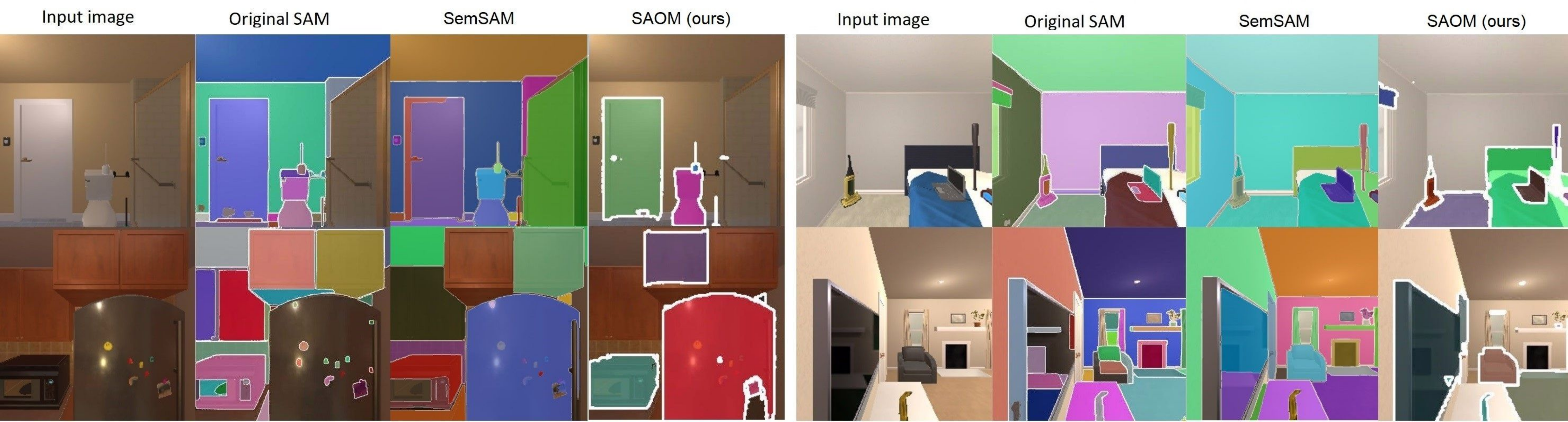}

   \caption{Comparison between vanilla SAM, Semantic-SAM (SemSAM) and our SAOM on images from real-to-sim test set, where we adopt the ``everything'' mode to obtain SAM segmentation for 4 different scene types. Note that objects such as bed, laptop and blinds in a bedroom, microwave and fridge in a kitchen, TV and armchair in a living room or toilet and plunger in a bathroom have a whole-object segmentation mask with SAOM.}
   \label{fig:sam11}
\vspace{-3mm}
\end{figure*}
We are not aware of any metrics used to evaluate the performance of SAM in the ``everything'' mode, as the model provides only segmentation masks for a given image without assigning labels to them. Hence, we propose to evaluate an object mask by segmentation IoU and accuracy scores from a single foreground point perspective, then combine those metrics per class. Since the original SAM can predict multiple masks for a single point, we evaluate only the model’s most confident mask for each object. 

The comparison of our SAOM and the original SAM is presented in Table \ref{tab:pick} for pickupable and receptacle objects of different sizes and in Table \ref{tab:open} for openable objects. Note that these tables include only objects with more then 500 occurrences in the validation set. For all 54 object classes SAOM has an increase in mIoU from 10.82 to 39.17 (28\% improvement) and in mAcc from 40.63 to 65.62 (25\% improvement) in comparison with SAM. The mean classification accuracy for SAOM is 0.36 across 54 object classes.

Tables \ref{tab:pick} and \ref{tab:open} show that all frequently-seen object classes have a significant increase in IoU scores in comparison with the vanilla SAM. Although, for some object classes the segmentation accuracy values decrease for SAOM (highlighted in red) (e.g CounterTop or Dresser), this is due to the challenging object appearance in our dataset. Receptacle objects tend to be heavily occluded, therefore an object mask can be split on several unconnected parts, making boundaries prediction difficult. As for openable objects, there are many examples in the dataset where these objects appear in the background of simulated scenes. Therefore, their size is relatively too small, which can decrease the segmentation accuracy score.

The qualitative results in the ``everything'' mode are presented in Fig. \ref{fig:sam11} for both SAOM and the original SAM model. We additionally compare SAOM with Semantic-SAM, as it is able to work in the ``everything'' mode and aims to output instance level (whole) object masks. In comparison with the original SAM and Semantic-SAM, SAOM tends to predict large and medium size objects with a whole-object segmentation mask. Most house general structures (such as walls and windows) and non-interactable objects (such as pictures and mirrors) are removed during predictions. Hence, SAOM can be considered to be working in the ``everything of interest'' mode, as the model targets 54 frequently-seen classes in indoor environments from the fine-tuning dataset. Please refer to the \href{https://sigport.org/documents/supplementary-materials-segment-any-object-model-saom}{supplementary material} for additional qualitative examples.

As a result of the ``everything of interest'' mode, SAOM significantly reduces the amount of the output segmentation masks. We calculated the amount of output masks by applying SAM, Semantic-SAM and SAOM on our Real-to-Sim validation set, containing 6500 images (Table \ref{tab:count}). In comparison with both Semantic-SAM and the original SAM, our SAOM model yields 39\% and 81.6\% fewer masks, respectively. SAM and Semantic-SAM tends to oversegment objects, focusing on parts and sub-parts rather than the entire objects. SAOM, on the other hand, is more object-centric, as indicated by the reduced number of masks. This distinction is quantified by higher IoU and Accuracy scores in Tables \ref{tab:pick} and \ref{tab:open}.

\subsection{Single-Object Segmentation}

\begin{figure}[t]
   \centering
 \vspace{-3mm}
   \includegraphics[width=0.98\linewidth]{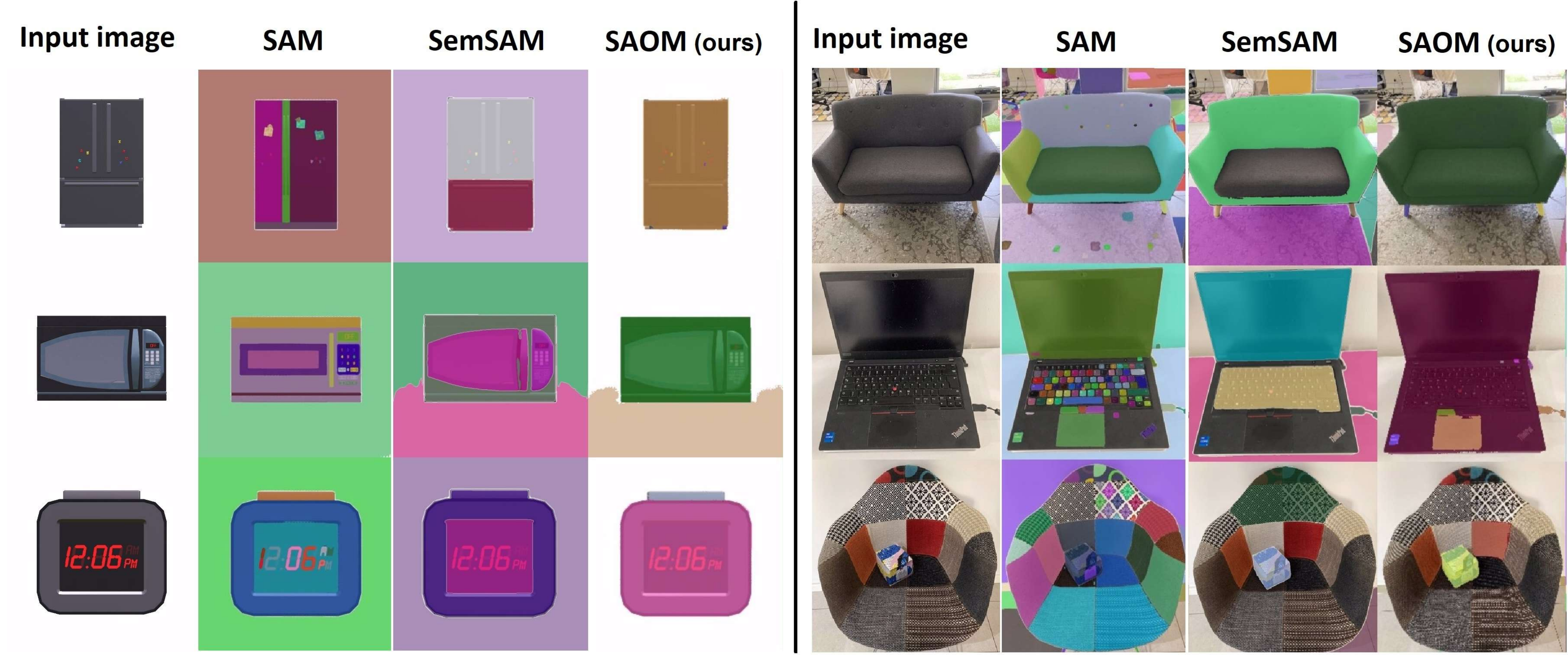}

   \caption{Comparison between Semantic-SAM (SemSAM), our SAOM and vanilla SAM on images from our single-object test set obtained from Ai2Thor simulator (left) and real indoor environments (right), where we adopt the ``everything'' mode to obtain segmentation masks for objects of various sizes. SAOM divides objects on less parts than vanilla SAM and Semantic-SAM in all test samples.}
   \label{fig:mix}
\end{figure}

We conducted an additional experiment to evaluate SAOM's generalization abilities for single object segmentation task in the ``everything'' mode. An additional test set of single object images was collected within Ai2Thor test scenes. The single-object segmentation results from Fig. \ref{fig:mix} (left side) confirm that SAOM performs better than SAM and Semantic-SAM for personalized object mask prediction. While Semantic-SAM tends to divide objects into a lesser number of parts in comparison with the original SAM, our SAOM received near perfect whole-object masks. Please refer to the \href{https://sigport.org/documents/supplementary-materials-segment-any-object-model-saom}{supplementary material} for additional qualitative examples.
\begin{figure}[t]
   \centering

   \includegraphics[width=0.98\linewidth]{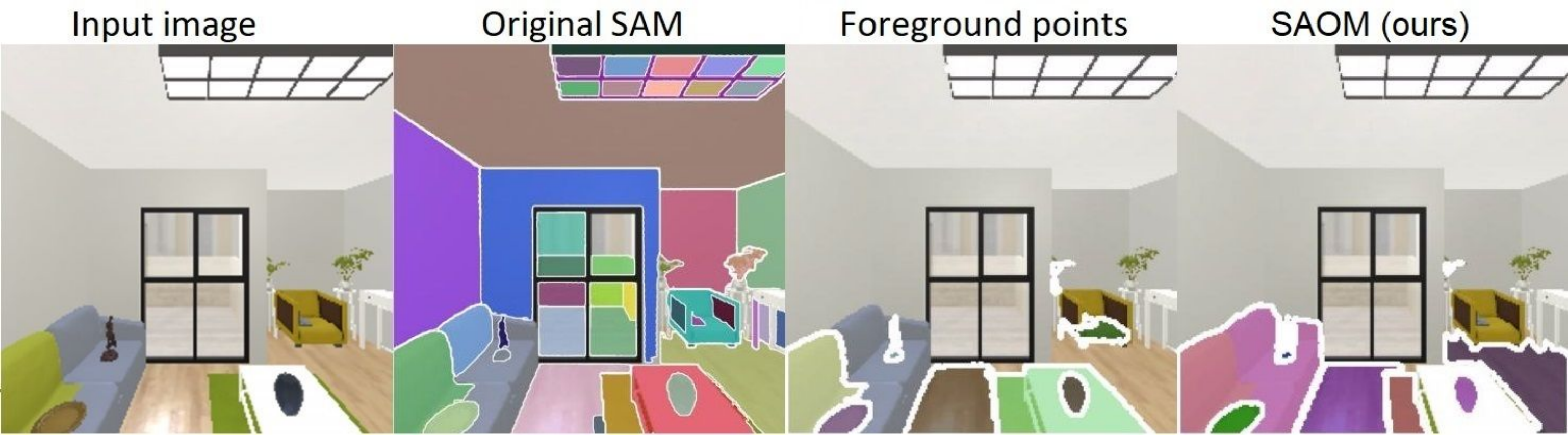}

   \caption{Comparison between vanilla SAM model, SAOM with simple foreground object points selection and SAOM with \textit{nearest neighbour assignment} method. We adopt  the ``everything'' mode to obtain segmentation masks. Note that sofa is segmented as a whole object only with SAOM.}
   \label{fig:sam15}
\end{figure}

\begin{table}[tp!]
\centering
\caption{Output Masks Count for the Real-to-Sim Val Set}
\label{tab:count}
\scalebox{0.90}{
\begin{tabular}{|l|l|l|l|}
\hline
                       & \cellcolor[HTML]{CBCEFB}SAM & \cellcolor[HTML]{CBCEFB}SemSAM & \cellcolor[HTML]{CBCEFB}SAOM \\ \hline
Number of output masks & 168691                        & 66280                       & 31015                        \\ \hline
\end{tabular}%
}
\vspace{-3mm}
\end{table}

\subsection{Selection of Object Point Prompts}
We conducted another experiment to investigate how the selection of point prompts affects the inference results in the ``everything'' mode. We compare in Fig. \ref{fig:sam15} the output results for: vanilla SAM, our SAOM (trained for 200 epochs) using the proposed \textit{nearest neighbour assignment} method and SAOM trained for 300 epochs using simple foreground object points (without \textit{nearest neighbour assignment}) similar to PerSAM [28]. All methods were trained using the SAM's standard 32 × 32 input point grid.

The experiments confirm the importance of our \textit{nearest neighbour assignment} method in the ``everything'' mode. While the model using simple foreground object points was trained for more epochs, it did not achieve the target result of the whole-object segmentation. 

\subsection{Sim-to-Real Inference Stage}
\label{sec:realdataset}
For the sim-to-real inference stage, we created an additional challenging test set with 151 images of objects in real life environments with various backgrounds (a box on a table or on a sofa), various view points (a box visible from one side or from the top), various states for openable objects (closed or opened laptop) and different variations of the same object.

We decided to use our own test set for the sim-to-real inference stage to exclude possible data bias with the SA-1B dataset, used by the original SAM model. Moreover, our sim-to-real test set includes both personalized object images and images with several objects at a time, as well as images of receptacle objects occluded with pickupable objects to test different levels of occlusion. Outputs of the vanilla SAM, Semantic-SAM and SAOM are shown in Fig. \ref{fig:mix} (right). 

Our experiments show that SAOM generalizes well to real-world data and can be trained without any real-world data samples to predict the the whole-object segmentation mask.
SAOM can predict high-quality masks for unoccluded or slightly occluded receptacle objects. On the other hand, if the receptacle object is mostly occluded, SAOM concentrates on the small pickupable objects on top of the receptacle object (an armchair in Fig. \ref{fig:mix}). For small pickupable objects, predictions depend on the object background. If the background is challenging, SAOM tends to focus on it instead of the foreground objects of interest. Please refer to the \href{https://sigport.org/documents/supplementary-materials-segment-any-object-model-saom}{supplementary material} for additional qualitative examples.

\section{LIMITATIONS}

One of the limitations of SAOM is that it tends to predict masks of the foreground objects better than for the background ones. This is due to the relatively small size of the background objects (Fig. \ref{fig:sam11}). To overcome this problem, we propose to use a point grid with larger number of points (e.g. 64 × 64 points instead of the standard 32 × 32). Another solution could be to use a stronger backbone model instead of ViT-B. The second limitation of SAOM lies in the low classification accuracy score, which should be further improved (e.g. by using data augmentation techniques).

\section{CONCLUSIONS}

This paper makes three contributions for a multi-class multi-instance segmentation task in interactive indoor environments. First, we study the possibility of direct transfer of SAM's fine-tuning strategy trained in simulated environments to the real world without using any real-world data during training.  In the real-to-sim training phase, we substitute the original object point prompts with their nearest neighbours from a pre-defined point grid on an image to make the model functional in the ``everything'' mode. 

Second, we present a fine-tuned version of SAM - SAOM - adapted for semantic segmentation of whole objects in indoor environments. The experimental results show that our SAOM can effectively and efficiently generalize for real-life scenes with low training costs. Moreover, we present a novel dataset collected from Ai2Thor simulator for the real-to-sim training stage. We evaluate our approach using classic segmentation IoU and accuracy metrics in a personalized manner and show that it outperforms the original SAM model after fine-tuning.

In future, we intend to combine our Real-Sim strategy with domain adaptation or domain randomization methods to enhance its generalization abilities on real life scenes.


\bibliographystyle{IEEEbib}
\bibliography{root}

\end{document}